\documentclass[sigconf]{acmart}

\usepackage{booktabs} 
\usepackage{multirow}
\usepackage{algpseudocode}
\usepackage[bottom]{footmisc}
\usepackage{blindtext}
\usepackage{graphicx}
\usepackage[font=footnotesize]{caption,subcaption}
\usepackage{wrapfig}

\usepackage{flushend}
\usepackage{amsmath,amsfonts,amsthm,bm}
\usepackage{stfloats}
\usepackage{graphicx}
\usepackage{textcomp}
\usepackage{xcolor}
\usepackage{booktabs} 

\usepackage{amsmath}
\usepackage{rotating}
\usepackage{booktabs}
\usepackage{listings}

\usepackage{listings}

\usepackage{ragged2e} 
\usepackage{algorithm}
\usepackage[algo2e]{algorithm2e} 

\AtBeginDocument{%
  \providecommand\BibTeX{{%
    \normalfont B\kern-0.5em{\scshape i\kern-0.25em b}\kern-0.8em\TeX}}}

\setcopyright{acmlicensed}

\copyrightyear{2024}
\acmYear{2024}
\setcopyright{acmcopyright}\acmConference[GLSVLSI '24]{Proceedings of the Great Lakes Symposium on VLSI 2024}{June 12--15, 2024}{Tampa, Florida, USA}
\acmBooktitle{Proceedings of the Great Lakes Symposium on VLSI 2024 (GLSVLSI '24), June 12--15, 2024, Tampa, Florida, USA}

\acmPrice{15.00}
\acmDOI{10.1145/3453688.3461760}
\acmISBN{978-1-4503-8393-6/21/06}

\begin{document}

\title{FFCL: Forward-Forward Net with Cortical Loops, Training and Inference on Edge Without Backpropagation}

\author{Ali Karkehabadi, Houman Homayoun, Avesta Sasan}
\affiliation{\vspace{0.1cm}
\institution{University of California, Davis, CA, USA.}\country{}
}
\email{{akarkehabadi, hhomayoun, asasan}@ucdavis.edu}

\begin{abstract}
The Forward-Forward Learning (FFL) algorithm is a recently proposed solution for training neural networks without needing memory-intensive backpropagation. During training, labels accompany input data, classifying them as positive or negative inputs. Each layer learns its response to these inputs independently. In this study, we enhance the FFL with the following contributions: 1) We optimize label processing by segregating label and feature forwarding between layers, enhancing learning performance. 2) By revising label integration, we enhance the inference process, reduce computational complexity, and improve performance. 3) We introduce feedback loops akin to cortical loops in the brain, where information cycles through and returns to earlier neurons, enabling layers to combine complex features from previous layers with lower-level features, enhancing learning efficiency.

\end{abstract}

\algdef{SE}[DOWHILE]{Do}{doWhile}{\algorithmicdo}[1]{\algorithmicwhile\ #1}%



\maketitle

\pagestyle{empty}

{\fontsize{8pt}{8pt} \selectfont
\textbf{ACM Reference Format:} \\
Ali Karkehabadi, Houman Homayoun, and Avesta Sasan.
2024. Forward-Forward Net with Cortical Loops:  A Backpropagation-Free Learning Solution for Training and Inference on Edge. In \it{GLSVLSI '24: Great Lakes Symposium on VLSI, June 12--15, 2024, Tampa, Florida, USA.} ACM, New York, NY, USA, 7 pages.}

\section{Introduction}
Deep learning has revolutionized problem-solving methodologies, with backpropagation serving as the cornerstone technology enabling the learning process. The backpropagation algorithm aims to fine-tune network parameters to minimize the discrepancy between the network's predictions and the actual ground truth\cite{ma2020hsic}. This process leverages gradient descent, utilizing the chain rule of differentiation to compute the loss function's gradient, allowing for the backward propagation of error. This mechanism enables updates to network parameters in a direction opposite to the gradient, optimizing network efficacy.

Over the past decade, backpropagation has undergone significant evolution, incorporating new features to address challenges encountered when training deep learning models, such as vanishing or exploding gradients. While these advancements have improved performance, backpropagation remains resource-intensive, demanding massive memory and computational power. Unlike the inference phase, where intermediate activations are consumed soon after they are generated, backpropagation necessitates the retention of all intermediate activations in memory for weight adjustment. This memory-intensive nature presents a considerable challenge, especially when deploying backpropagation on edge devices with limited resources. In such settings, the constraints on memory and compute capabilities exacerbate the difficulty of implementing backpropagation effectively, hindering its practical application in resource-constrained environments. On the other hand, Deep Learning (DL) systems, though inspired by the intricate structure and functionality of the human brain, diverge considerably in their training mechanisms. Notably, there is a lack of concrete evidence to suggest that the brain employs a learning methodology akin to the backpropagation algorithm used in training DL models \cite{lillicrap2020backpropagation};   The BP algorithm adjusts the weights of connections between artificial neurons based on the gradient of the error function, a process not directly observed in biological neural networks. The brain's ability to efficiently learn and generalize from a limited set of examples far surpasses current DL models \cite{zador2019critique}. This efficiency hints at a learning paradigm in the brain that is fundamentally different from the iterative error reduction in machine learning. Neuroscientific research indicates the existence of feedback mechanisms within the brain, characterized by complex neural loops rather than the straightforward, error-correcting backpropagation paths found in artificial neural networks \cite{forsyth1999object}. These loops suggest a more dynamic and possibly more efficient information processing system, where forward and backward signals may contribute to learning in a manner not fully replicated by current DL training methodologies.

Recognizing the substantial memory demands of conventional training methods and the limitations of implementing such methods on edge devices, alongside the brain's capability to learn without relying on backpropagation, Geoffrey Hinton introduced the concept of FFL \cite{hinton2022forward}. This innovative approach aims to markedly diminish the memory requirements for training at the edge, bringing them down to levels comparable to those needed for inference with no need for storing the activations. This methodology not only offers a more memory-efficient solution for edge computing but also aligns more closely with our current understanding of neural learning processes in the brain. However, it introduces a complexity in the inference phase, necessitating the separate evaluation of each potential outcome by concurrently inputting the data and respective labels. 
In our research, we delve into the practicality and efficacy of FFL, contributing several enhancements that support this emerging field and enhance the FFL solution. Our findings are intended to stimulate broader interest and participation in this promising area of study. Specifically, we have developed refinements to the FFL algorithm that 1) enhance its learning capabilities, and 2) substantially decrease the computational burden during the inference stage. Through sharing our insights and improvements, we aim to encourage wider adoption and further exploration of this novel approach within the research community.

\vspace{-10pt}
\section{Background}
Predictive Coding (PC) \cite{aitchison2017or} is a theory suggesting the brain predicts sensory input based on past experiences, focusing on discrepancies between predictions and actual input to efficiently process information. This framework explains how perceptions are formed, facilitates learning through error correction, and provides insights into various neurological and psychiatric conditions by highlighting the brain's active role in shaping our sensory experiences and responses to the environment. The FFL algorithm, motivating this work, attempts to emulate the brain's learning processes and in this context adheres to predictive coding. In the PC theory, the brain is considered a predictive system where each layer strives to enhance the accuracy of its own inputs. In this context, as prior art to FFL, we overview the various forms of supervised predictive coding solutions investigated in DL. In a novel approach presented by Dellaferrera et al. \cite{dellaferrera2022error}, the backpropagation was replaced with a dual forward-pass solution. This second pass adjusts the input based on network errors, effectively mitigating the need for symmetric weights, removing the reliance on distant learning signals, and the cessation of neural activity during error backpropagation. Further expanding the landscape of neural network training, Kirsch et al. \cite{kirsch2021meta} introduced the Variable Shared Meta Learning (VSML) framework that replaces backpropagation by only forward operations. This algorithm harmonizes various meta-learning methodologies, showcasing that through weight-sharing and benefiting from network sparsity sophisticated learning could be expressed. In another innovative stride, Baydin et al. \cite{baydin2022gradients} proposed a technique, denoted as forward gradient, for computing gradients using directional derivatives that are obtained in forward-mode differentiation. The final work, presented in this paper as background, is the solution proposed by Geoffery Hinton, denoted as Forward-Forward learning. Given this paper extends the FFL algorithm, in this section we dive deeper into this concept and cover greater details:

Introduced by Geoffrey Hinton, the Forward-Forward algorithm presents a novel paradigm in neural network training by substituting the conventional backpropagation technique with an additional forward pass. Within this framework, the initial pass is tasked with generating predictions, while the subsequent forward pass is dedicated to refining the model in light of errors detected during the initial prediction phase. This approach endeavors to surmount several of backpropagation's constraints, including its dependency on symmetric weights and the backward error transmission, by confining all modifications to forward operations, thus aligning more closely with mechanisms of biological learning.

In this algorithm, the positive or initial pass processes genuine data, adjusting weights to enhance a defined 'goodness' measure across each layer. In contrast, the negative pass utilizes "negative data" to reduce this 'goodness' measure. Investigating two particular 'goodness' metrics—the sum of squared neural activations and their reciprocal—highlights the potential for adopting diverse metrics in this context.

The learning mechanism, as described in Equation \ref{eq1}, is designed to ensure the 'goodness' measure for authentic data substantially surpasses a predetermined threshold, while remaining markedly below this threshold for negative data. The algorithm aims to accurately classify input vectors as either positive or negative, estimating the likelihood of input being positive by applying the logistic function, $\sigma$, to the 'goodness' metric, offset by a threshold, $\theta$. This innovative approach signifies a leap towards enhancing the efficiency and biological fidelity of neural network training, opening new pathways for advancements in machine learning research and its practical deployments.
\vspace{-3pt}
\begin{equation}
\small
G(\text{X}) = \sigma((\sum_{j} (x_{j}^2) - \theta))
\label{eq1}
\end{equation}

This approach underscores a strategic pivot towards enhancing the efficiency and applicability of neural network training, eschewing conventional learning paradigms.

\begin{figure}[hbt!]
    \centering
    \includegraphics[width=0.47\columnwidth]{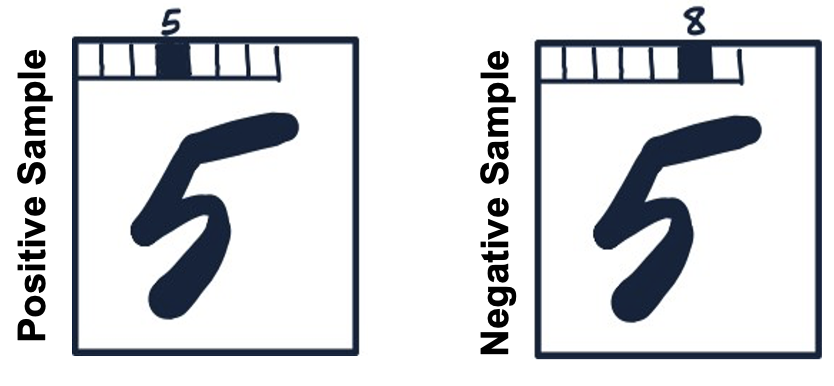}
    \vspace{-12pt}
    \caption{\cite{hinton2022forward} embeds labels in MNIST's black border, altering the first 10 pixels for class representation: '1' for the true class in positive samples and '1' in a random other class for negative samples, with the rest as '0'}
    \label{Mnist}
\end{figure}

Algorithm \ref{alg} captures the FFL procedure. The algorithm iteratively adjusts the model's weights over a predefined number of epochs and iterations within each epoch. In each iteration, the algorithm identifies positive samples, where the class is correctly labeled, and negative samples, which are randomly chosen from non-matching classes. For both sets of samples, it computes the activations at each layer of the model.  Fig. \ref{Mnist} illustrates the technique for generating negative and positive data in the forward-forward algorithm. The algorithm computes the gradients of the loss function, which is designed to penalize the model for incorrect classifications. Specifically, for positive samples (\(g_{\text{pos}}\)), the loss increases when the model's confidence in the correct classification is low. For negative samples (\(g_{\text{neg}}\)), the loss increases when the model incorrectly classifies them as positive. 
In each epoch of the Forward-Forward Algorithm, both positive and negative data are utilized to calculate the loss function. This loss is computed as the sum of the differences between a pre-defined threshold $\theta$ and the 'goodness' measure $G$, for both positive $x_{\text{pos}}$ and negative $x_{\text{neg}}$ inputs. The 'goodness' measure $G(x)$ for any input $x$ is obtained by applying the ReLU activation function to the matrix multiplication of $x$ with the transposed weight matrix plus a bias term:
\begin{equation}
\small
G(x) = \text{ReLU}(X\times W^{T} + b)
\end{equation}

The loss function can be defined as:
\begin{equation}
\small
\text{Loss} = (\theta - G(x_{\text{pos}})) + (G(x_{\text{neg}}) - \theta) = - G(x_{\text{pos}}) + G(x_{\text{neg}})
\end{equation}
This loss aims to optimize the 'goodness' measure by maximizing it for positive inputs and minimizing for negative inputs relative to the threshold $\theta$.
So The loss function is defined as:

\begin{equation}
\small
loss = (\log(1 + e^{-g_{pos}}) + \log(1 + e^{g_{neg}})) / 2    
\end{equation}

This loss function combines the penalties for both types of errors in a manner that encourages the model to correctly classify both positive and negative samples with high confidence. Through single-layer BP (and not full BP), the algorithm updates the model's weights according to gradients, progressively reducing the classification error over time. To generate positive and negative samples for the Forward-Forward Algorithm, the algorithm modifies the first N (the number of classes) pixels.





\begin{algorithm}[H]
\RaggedRight
\SetAlgoLined
\caption{Forward-Forward Algorithm }
\For{$\text{l} \in \text{model.layers}$}{
    \For{$\text{e} \in \text{MaxEpock}$}{
            \# \textcolor{blue}{Prepare pos and neg samples}\\
            $x_{e}, L_{e+} = get\_training\_sample(e)$\\            
            $L_{pos}, L_{neg} = \{C\{0\}\};$       \# concat C 0s   \\
            $L_{e-} = random(0, C, \! L_{e+})$;\\
            $L_{pos}[L_e]=1;$ \# change 0 in label position to 1 \\
            $L_{neg}[L_{ne}]=1;$ \# change 0 in a non label position to 1\\
            $x_{pos} = replace\_boarder(x_e, L_{pos})$ \\
            $x_{neg} = replace\_boarder(x_e, L_{neg})$ \\

            \# \textcolor{blue}{Run pos and neg samples to target layer}\\
            $g_{\text{pos}} = \text{RunLayers}(0, l, x_{pos})$ \\
            $g_{\text{neg}} = \text{RunLayers}(0, l, x_{neg})$ \\
            \# \textcolor{blue}{Compute loss}\\
            $\text{loss} = \frac{1}{2}(\log(1 + e^{-g_{\text{pos}}}) + \log(1 + e^{g_{\text{neg}}}))$ \\
            \# \textcolor{blue}{Update weights}\\
            $W_{grad} = \text{one\_layer\_backpropagate}(loss);$ \\
            $\text{model.layer(l).weights\_update}(W_{grad});$
    }
}
\label{alg}
\end{algorithm}

To construct positive samples, the algorithm marks the tensor index for the target class as 1, setting all others to 0, thus denoting the class's presence. Conversely, for negative samples, it chooses a random index, not of the actual class, to mark as 1, leaving the rest at 0, to indicate the class's absence. This method allows clear differentiation between classes by specifying class presence in positive samples and absence in negative ones. For example, for the MNIST dataset with 10 classes, this differentiation is achieved by altering the initial 10 pixels to represent class information.

During training, each layer is trained for a specified number of epochs before proceeding to the next layer. This differs from the traditional multilayer perceptron (MLP) approach, where all layers are typically trained simultaneously. In the Forward-Forward Algorithm, each layer learns independently and is only influenced by the outputs of the preceding layers, enhancing the specificity and efficiency of the learning process.

\begin{figure}[hbt!]
    \centering
    \includegraphics[width=0.99\columnwidth]{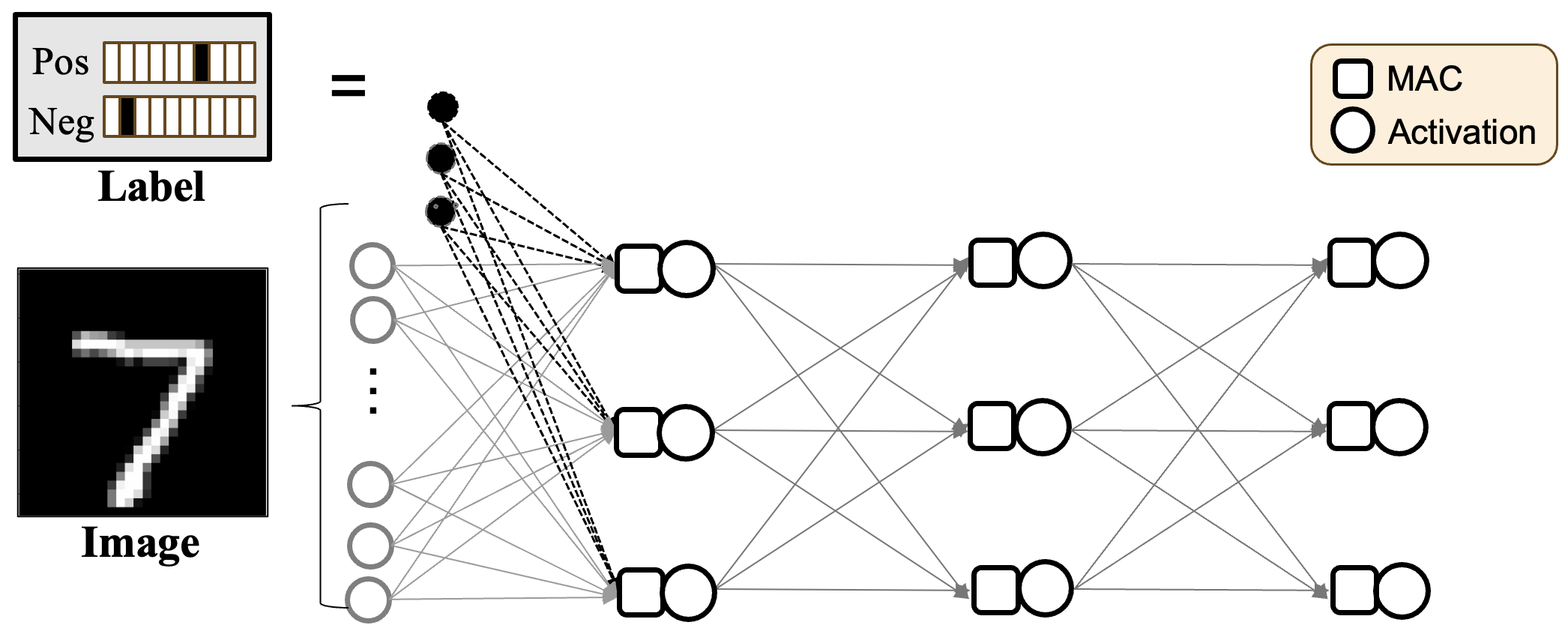}
    \vspace{-5pt}
    \caption{Forward-Forward Algorithm Setup: Transforms MNIST images to a 784-unit input layer. For class representation, the initial N pixels are adjusted: '1' for the correct class in positive samples and '1' in a non-class index for negative samples, particularly in the first 10 pixels for MNIST. Layers are trained sequentially.}
    \label{ffregular}
    \vspace{-10pt}
\end{figure}

\subsection{Motivation and Problem Statement}
In the Forward-Forward Learning (FFL) approach, we encountered several challenges during training and inference. Initially, we found that input labels were directly provided only to the first layer, with subsequent layers receiving a blend of label and feature information. We theorized that this method of indirect labeling, which merges data and features, complicates the learning process for layers beyond the first. As a consequence of this labeling strategy, the inference phase's computational demand is increased by the number of classes, necessitating separate computations for each class by inputting its label into the first layer and processing through subsequent layers. For instance, in a 10-class scenario, this would require executing the network 10 separate times. While FFL mitigates memory constraints during training, it introduces heightened computational complexity during inference. Therefore, we explored alternative methods that could significantly reduce computational demands at inference. Moreover, although FFL draws inspiration from the hypothesis of the brain's Contrastive Excitatory (CE) learning process, it maintains a strictly forward flow of information within a Directed Acyclic Graph (DAG) structure. We hypothesize that incorporating feedback mechanisms, while retaining FFL's training efficiencies, could enhance the model's learning capabilities, albeit at the cost of increased training complexity since each node might undergo training during both the forward pass and potentially multiple rounds of feedback. Nonetheless, we anticipate that training incorporating feedback mechanisms, following the initial feedforward network training, could be executed more swiftly. Consequently, we propose a methodology designed to augment FFL's learning precision while simultaneously curbing computational burdens during inference.


\section{Methodology}
We introduce a Feed Forward with Cortical Loop (FFCL) approach that builds on FFL, offering improvements through direct labeling of each hidden layer and incorporating feedback loops into the model architecture, along with a revised training methodology. The first enhancement boosts training accuracy, shortens training duration, and reduces inference complexity. The second adjustment further reduces the model's complexity with a marginal increase in model complexity. We detail our two-phase solution as follows:

\subsection{Direct Label Feeding}
Our methodology enhances supervised FFL training by directly inputting label information into each layer, diverging from FFL that blends and propagates label and feature data from the initial layer onwards. In the context of training, this approach involves appending the label to the input as additional data, effectively as padding, ensuring the integrity of the primary input remains intact. This unique strategy allows for distinct computations corresponding to positive and negative labels, which are then integrated with the computations arising from the interaction of input features and layer weights, plus the bias, before being passed through the activation function. The clear separation of label influence in the computational process enhances the specificity of information each layer receives, making the learning process more efficient.

Once the first hidden layer is trained, the subsequent layers employ only the outputs derived from the combination of input features and layer weights from the previous layer, along with the bias, excluding the direct label influence from further computations. This selective utilization of output streamlines the training process for subsequent layers, as it circumvents the need for managing distinct datasets for positive and negative labels, thereby simplifying the computational framework.

\begin{figure}[hbt!]
\vspace{-5pt}
    \centering
    \includegraphics[width=0.97 \columnwidth]{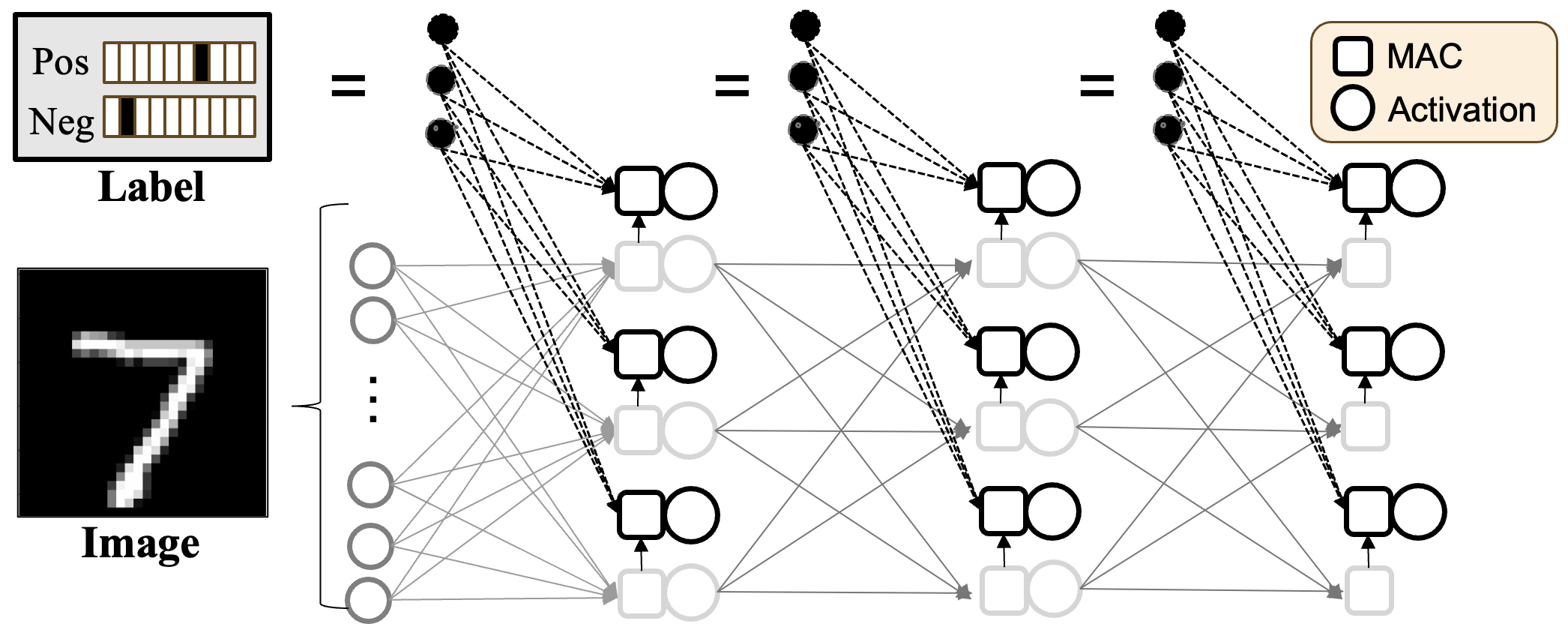}
    \vspace{-10pt}
    \caption{In our revised FFL we directly integrate class labels into the structure, maintaining original image integrity. Training links input and labels to initial layer neurons, enabling specialized computations. The following layers focus on weighted images and bias from the prior layer, optimizing processing. }
    \label{ff_new1}
    \vspace{-10pt}
\end{figure}

This streamlined approach extends through all subsequent layers, maintaining the clarity and specificity of label information without the computational overhead of separating positive and negative label datasets. Such an architectural innovation substantially mitigates the challenges associated with diluting label clarity across layers and the inefficiencies tied to training with distinct datasets for different label outcomes. Moreover, this methodology significantly diminishes the computational burden during the inference phase. By obviating the need for extensive class-based recalculations, the model can swiftly execute inference tasks by applying the relevant class labels to the learned features from the preceding layer, thereby enhancing the model's performance and efficiency.  Our approach, illustrated in Fig. \ref{ff_new1}, captures this advancement, demonstrating how each layer is directly labeled, and how features are directly propagated. 

In our solution, instead of replacing the boundary pixels of the image with the label pixels and using the label only in the computation of the first layer, we feed the label separately to each layer. The operation of each neuron, in Fig. \ref{ff_new1} is divided into MAC operation followed by activation. As shown, in Fig. \ref{ff_new1}, the neurons we use for computing the estimation of goodness (in dark black) and neurons used for propagating features to the next layer are separated. The labels only impact the estimation neurons. The estimation neurons are fed by the propagation neurons as a bias (1 input), and N weights for N labels. or more specifically, as shown in eq. \ref{mac_sum}, for each layer first we generate the multiplication and accumulation sum $s$ for each neuron using model weights connected to the input or previous layer activation $w_{in}$ and bias, then using eq. \ref{g_pos_calc} we compute the goodness, by using weights connected to labels $w_{label}$ and using $s$ as bias, and finally we prepare the activation of the neuron as input for the next layer in eq. \ref{activation_calc}. The complete detail of our algorithm is given in Algorithm \ref{alg_label_feed}.

\vspace{-8pt}
\begin{equation} \label{mac_sum}
     s = x \times w_{in}^T + b
     \vspace{-8pt}
\end{equation}

\begin{equation} \label{g_pos_calc}
     g_{pos/neg} = Relu(Label \times w_{label}^T + s)
     \vspace{-8pt}
\end{equation}

\begin{equation}\label{activation_calc}
     a = Relu(s)
     \vspace{-8pt}
\end{equation}

\begin{algorithm}[H]
\RaggedRight
\SetAlgoLined
\caption{Forward-Forward with Direct Label Feed (FFDL) }
\For{$\text{l} \in \text{model.layers}$}{
    \For{$\text{e} \in \text{MaxEpock}$}{
            \# \textcolor{blue}{Get new training sample}\\
            $x_{e}, L_{e+} = get\_training\_sample(e)$\\     
            \# \textcolor{blue}{use the input to run to target layer}\\
            $x_e = \text{RunLayer}(0, l, x_e);$ \#Activation output stored in $x_e$\\
            \# \textcolor{blue}{Prepare pos and neg samples}\\
            $L_{pos}, L_{neg} = \{C\{0\}\};$       \# concat C 0s   \\  
            $L_{e-} = random(0, C, \! L_{e+})$;\\
            $L_{pos}[L_e]=1;$ \# change 0 in label position to 1 \\
            $L_{neg}[L_{ne}]=1;$ \# change 0 in a non label position to 1\\
            $x_{pos} = \{x_e, L_{pos}\}$ \# concat with label \\
            $x_{neg} = \{x_e, L_{neg}\}$ \# concat with label \\   
            \# \textcolor{blue}{Run pos and neg samples to target layer}\\
            $g_{\text{pos}} = \text{RunLayers}(0, l, x_{pos})$ \\
            $g_{\text{neg}} = \text{RunLayers}(0, l, x_{neg})$ \\
            \# \textcolor{blue}{Compute loss}\\
            $\text{loss} = \frac{1}{2}(\log(1 + e^{-g_{\text{pos}}}) + \log(1 + e^{g_{\text{neg}}}))$ \\
            \# \textcolor{blue}{Update weights}\\
            $W_{grad} = \text{one\_layer\_backpropagate}(loss);$ \\
            $\text{model.layer(l).weights\_update}(W_{grad});$
    }
}
\label{alg_label_feed}
\end{algorithm}





            



 As mentioned earlier, this direct label feeding approach not only improves accuracy but also reduces the inference computational cost. We will review the impact of direct label feed in the result section of this work, however, next we elaborate on how this approach reduces computational complexity. 
The original FFL is an MLP. Let's assume an MLP with M input neurons and N output neurons. The number of required FLOPS to compute each layer in the FFL is computed as $MN + 2N$, where the added $2N$ accounts for the cost of activation and also the addition of the bias term. However, note that in the original FFL, to perform the classification, each of the labels has to be fed to the first layer, and since the addition of the labels changes the activations in each layer, all layers need to be recomputed. Hence, for C classes, each layer needs to be computed C times, resulting in:
\begin{equation}\label{FlopcountFFL}
\small
\text{FFL FLOPs/Layer = } MNC + 2NC 
\end{equation}

In our solution, however, the computation of each layer is done with label-independent activation from the previous layer and the application of new labels. For computing the label-independent neurons, we have $NM + N$ operations. For computing the goodness for one label we have $CN + N$ operation, where C is the number of labels, and N accounts for the addition of sum as bias. Finally, for the computation of all labels (C of them), we have $C^2N+CN$ operations. 
In the result, the total number of operations for each layer to measure the goodness across all classes is:
\vspace{-3pt}
\begin{equation}\label{FlopcountFFDL}
\small
\text{FLOPs/Layer = } MN + NC^2 + NC  + 2N
\end{equation}

Removing N from both equations, the advantage comes when the following inequality holds: 
\vspace{-5pt}
\begin{equation}
\small
 M + C^2 + 2 < MC + C 
\end{equation}

which could, with some estimation simplified as advantagous when $M>C+1$. With MNIST, C is 10, and on average M is 500 in the model used in FFL.

\subsection{Forward-Forward Net with Cortical Loops}

\begin{figure*}[hbt!]
    \centering
    \includegraphics[width=1.95\columnwidth]{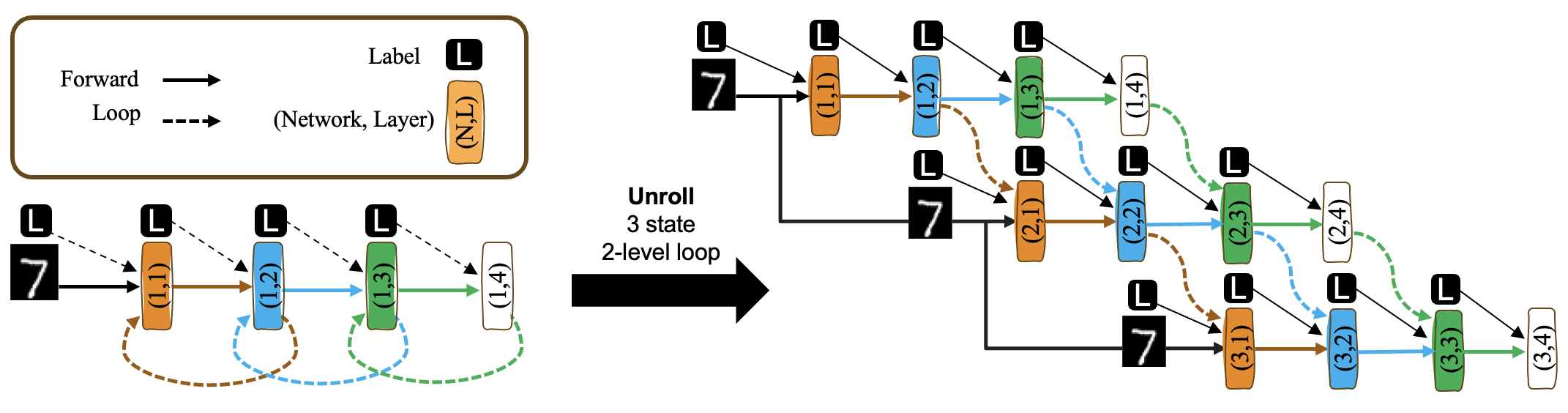}
    \vspace{-13pt}
    \caption{Architecture of Forward-Forward Net with Cortical Loops. To train the model effectively using existing training engines, the model is unrolled. The number of times, the model is unrolled decides the degree to which feedback information propagates in the system. For example, with a 3 (N) layer network, each layer feedback could reach 2 (N-1) previous layers.}
    \label{back}
    \vspace{-12pt}
\end{figure*}

In our second key contribution, we incorporate feedback mechanisms analogous to those found in the brain, enabling each layer to provide feedback to the preceding one. Given the complexities involved in training models with feedback, we employ a technique that involves unrolling the network multiple times and sharing weights across these unrolled instances. The extent of unrolling determines how far back the information can propagate through the network. For instance, duplicating the network twice, as depicted in Fig. \ref{back}, allows information from each layer to influence up to two preceding layers. In this setup, a weight-sharing strategy is adopted to ensure consistency across all duplicates of the network, meaning that updating the feedforward weights in the first layer concurrently updates those in the subsequent duplicates.

This unrolling approach transforms the challenge of training a network with feedback into training an expanded feedforward network, which is readily manageable with existing learning frameworks. In our model, the same input is fed into each unrolled instance, simulating the effect of feedback on a static image. This technique could also accommodate variations of the input image, such as augmented or rotated versions, potentially enhancing model robustness. Furthermore, this architecture could facilitate the processing of time-sequenced images from the same scene, offering rich temporal information that might enhance learning. While these applications remain unexplored in the current work, they underscore the potential of our feedback-based architecture, which we aim to investigate in future studies.

The proposed unrolling approach also introduces considerations regarding training schedules. Training can occur for the feedforward path as soon as the previous layer is ready, and for the feedback path when the subsequent layer is trained. Various training schedules are conceivable, ranging from multiple forward passes capturing information from subsequent layers as it becomes available, to Just-In-Time training that initiates feedback loop training as soon as it's feasible. Although exploring the impact of different training schedules is beyond the scope of this paper and reserved for future research, we present a single training methodology without asserting its superiority. It's important to note that the model's complexity is only marginally increased by the added feedback mechanism, thanks to weight sharing, thus preventing significant growth in model size. For inference, we evaluate the accuracy of the layers in the final unrolled instance, which contains the bulk of the feedback information. Accuracy metrics are reported both individually for each layer and collectively for the entire model, with the latter being the aggregation of positive and negative votes for each class across all layers in the final instance. Unlike a simple average, this ensemble approach combines votes, offering a nuanced measure of accuracy. The training algorithm for our unrolled model is detailed in Algorithm \ref{alg_Cortical_Loops}.

\begin{algorithm}[H]
\RaggedRight
\SetAlgoLined
\caption{Forward-Forward Net with Cortical Loops (FFCL)}
\For{$\text{n } \& \text{ l} \in \text{model.networks.layers}$}{
    \For{$\text{e} \in \text{MaxEpock}$}{
            \#\textcolor{blue}{In the sequence from 0 up to the $n + l$, } 
            $\textcolor{blue}{priority: 
            n} \leq \textcolor{blue}{l.}$\\
            \# \textcolor{blue}{Get new training sample}\\
            $x_{e}, L_{e+} = get\_training\_sample(e)$\\     
            \# \textcolor{blue}{use the input to run to target layer}\\
            $x_e = \text{RunLayer}(0, n, l, x_e);$ \#Activation output stored in $x_e$\\
            \# \textcolor{blue}{Prepare pos and neg samples}\\
            $L_{pos}, L_{neg} = \{C\{0\}\};$       \# concat C 0s   \\  
            $L_{e-} = random(0, C, \! L_{e+})$;\\
            $L_{pos}[L_e]=1;$ \# change 0 in label position to 1 \\
            $L_{neg}[L_{ne}]=1;$ \# change 0 in a non label position to 1\\
            $x_{pos} = \{x_e, L_{pos}\}$ \# concat with label \\
            $x_{neg} = \{x_e, L_{neg}\}$ \# concat with label \\
            \# \textcolor{blue}{Run pos and neg samples to target layer}\\
            $g_{\text{pos}} = \text{RunLayers}(0, n, l, x_{pos})$ \\
            $g_{\text{neg}} = \text{RunLayers}(0, n, l, x_{neg})$ \\
            \If{$\text{network != 0}$ }{
                \# \textcolor{blue}{Run pos and neg samples to target layer}\\
                $g_{\text{pos}}$ = $g_{\text{pos}}$ + $ \text{RunLayers}(0, n-1, l+1, x_{pos})$ \\
                $g_{\text{neg}}$ = $g_{\text{neg}}$ + $\text{RunLayers}(0, n-1, l+1, x_{neg})$ \\
                }
            
            \# \textcolor{blue}{Compute loss}\\
            $\text{loss} = \frac{1}{2}(\log(1 + e^{-g_{\text{pos}}}) + \log(1 + e^{g_{\text{neg}}}))$ \\
            \# \textcolor{blue}{Update weights}\\
            $W_{grad} = \text{one\_layer\_backpropagate}(loss);$ \\
            $\text{model.layer(l).weights\_update}(W_{grad});$
                
            }
    }

\label{alg_Cortical_Loops}

\end{algorithm}
\vspace{-10pt}

In Algorithm \ref{alg_Cortical_Loops}, we introduce Forward-Forward Net with Cortical Loops that employs a Parallel Direct Label Feeding mechanism, defined within an MLP (Multilayer Perceptron) framework. Here, `network` denotes the number of parallel instances of the Direct Label Feeding model, while `layer` refers to the number of layers within each network instance.

For networks identified by an index of 1, indicating the first network, there is an absence of backward input connections. Hence, the computation of \(g_{\text{pos}}\) and \(g_{\text{neg}}\) is exclusively based on the outputs from the last layer of this network. In contrast, for networks with indices greater than 1, the model incorporates inputs from two sources for the calculation of \(g_{\text{pos}}\) and \(g_{\text{neg}}\): outputs from the preceding layer and backward inputs from an adjacent layer. This approach is critical for the accurate calculation of the loss.

It is crucial to note the sequence of training for the layers in this model. The training process prioritizes the aggregate of the number of networks and the number of layers, under the stipulation that the network index is less than or equal to that of the layer index. This prioritization scheme facilitates a systematic and effective training regimen for the Parallel Direct Label Feeding model.

In our proposed architecture, feedback is currently limited to the immediately preceding layer and is applied whenever feasible, utilizing distinct weights from those in the feedforward path. This implementation represents just one approach to integrating feedback mechanisms. The potential exists for feedback to extend beyond the adjacent layer, creating additional loops within the network. Moreover, it's conceivable that not all feedback connections are necessary; a more sparse feedback architecture might be sufficient for our objectives, a topic of future study.


\section{Results}

Our evaluation of the enhanced FFL solution, which incorporates direct label input and cortical loops, was conducted on three benchmark datasets: MNIST \cite{lecun2010mnist}, Fashion-MNIST \cite{xiao2017fashion}, and CIFAR-10 \cite{krizhevsky2009learning}. Both MNIST and Fashion-MNIST datasets consist of images with dimensions of $28 \times 28$ pixels. These images were converted into a flattened format to form an input layer with 784 units. For the CIFAR-10 dataset, which contains images of size $3 \times 32 \times 32$, we processed the images in a similar fashion, resulting in an input layer of 3072 units. All computational experiments were performed using a NVIDIA T4 GPU, equipped with 2,560 CUDA cores, 320 Tensor cores, and 16 GB of GDDR6 memory. The models were trained and tested using FP16 precision. In each of the three experiments, a 4-layer MLP architecture was employed. The specific model architecture used in these experiments is detailed in Table \ref{model_architecture_table}.

\begin{table}[hbt!]
\centering
\scalebox{0.9}{
\begin{tabular}{lccc}
\hline
\textbf{Model Size}        & \textbf{MNIST} & \textbf{Fashion-MNIST} & \textbf{CIFAR10}\\ \hline                        
input  & 784                                & 784 & 3072                         \\

1st Layer  & 500                                & 500 & 2048                         \\
2nd Layer & 500                                & 500  & 1024                     \\
3rd Layer & 500                                & 500 & 512                        \\
4th Layer  & 500                               & 500 & 512                        
    
\\ \hline
\end{tabular}
}
\caption{The MLP model architecture used for training the FFL, FFDL, and FFCL for each selected dataset.}
\label{model_architecture_table}
\vspace{-20pt}
\end{table}

Each model underwent training 50 times, with each training session spanning 5,000 epochs using the Adam optimizer. We documented the accuracy for each layer individually and then calculated the overall model accuracy. The accuracies reported are averages from the 50 training iterations. Please note that the overall model accuracy is not simply an average of the layer accuracies. In line with the methodology established by FFL \cite{hinton2022forward}, model accuracy is determined by aggregating the 'goodness' votes from all layers before making a final decision. Thus, the model accuracy reflects the collective decision derived from the sum of votes, rather than an average of individual model layer decisions. Table \ref{results1_mnist} shows the accuracy comparisons of FFL, FFDL, and FFCL when trained on the MNIST dataset. The table clearly shows that each layer's accuracy in FFDL improves over FFL, and similarly, each layer's accuracy in FFCL improves over FFL. It is also noteworthy that while there is a significant drop in layer accuracies within FFL, the direct label feeding minimizes this drop in both FFDL and FFCL. Finally, the table highlights that the final model accuracy has significantly improved in both FFDL and FFCL, with FFCL achieving an accuracy of 97.23\% compared to FFL's 94.86\%.

\begin{table}[hbt!]
\centering
\scalebox{0.9}{
\begin{tabular}{lccc}
\hline
\textbf{Accuracy}        & \textbf{FFL\cite{hinton2022forward}} & \textbf{FFDL (Label Feed)} & \textbf{FFCL (Cortical Loop)}\\ \hline                        
1st Layer  & 94.77\%                                & 96.10\% & 96.72\%                         \\
2nd Layer & 94.43\%                                & 95.54\%  & 96.64\%                     \\
3rd Layer & 94.39\%                                & 94.93\% & 95.76\%                        \\
4th Layer  & 91.83\%                                & 94.77\% & 95.46\%                        \\
Model     & 94.86\% & 96.57\%                               & 97.23\%                                          
\\ \hline
\end{tabular}
}
\caption{Comparative results of the Forward-Forward method and the New Ideas for MNIST dataset}
\label{results1_mnist}
\vspace{-20pt}
\end{table}

Table \ref{results1_fashionmnist} presents the accuracy results of models trained on the FashionMNIST dataset. Due to the greater complexity of the FashionMNIST compared to MNIST, the accuracies reported are somewhat lower. Nonetheless, the enhancements from utilizing FFDL and FFCL are more distinct. The model accuracy of FFL was recorded at 86.04\%, whereas FFDL achieved a model accuracy of 87.6\% and FFCL reached 89.7\% accuracy, marking an improvement of nearly 4\% over FFL.

\begin{table}[hbt!]
\centering
\scalebox{0.9}{
\begin{tabular}{lccc}
\hline
\textbf{Accuracy}        & \textbf{FFL\cite{hinton2022forward}} & \textbf{FFDL (Label Feed)} & \textbf{FFCL (Cortical Loop)}\\ \hline                        
1st Layer  & 85.83\%                                & 86.11\% & 86.19\%                         \\
2nd Layer & 83.93\%                                & 84.21\%  & 84.64\%                     \\
3rd Layer & 82.81\%                                & 83.93\% & 84.12\%                        \\
4th Layer  & 81.19\%                                & 82.49\% & 83.98\%                        \\
Model     & 86.04\% & 87.61\%                               & 89.71\%                               
\\ \hline
\end{tabular}
}
\caption{Comparative results of the Forward-Forward method and the New Ideas for FashionMNIST dataset}
\label{results1_fashionmnist}
\vspace{-20pt}
\end{table}

Table \ref{results1_fashionmnist} documents the training accuracy of three models on the \ref{cifar10} dataset, which is the most complex dataset in our experiment. A very similar pattern is noted here. In FFL, the model accuracy significantly decreases in deeper layers, whereas in FFDL and FFCL, this drop is much less pronounced. Additionally, FFCL exhibits higher accuracy compared to FFDL, and FFDL shows higher accuracy compared to FFL across all layers and in overall model accuracy. As demonstrated, both the direct feeding of labels and the introduction of cortical loops effectively enhance model accuracy in forward-forward training.

\begin{table}[hbt!]
\centering
\scalebox{0.9}{
\begin{tabular}{lccc}
\hline
\textbf{Accuracy}        & \textbf{FFL\cite{hinton2022forward}} & \textbf{FFDL (Label Feed)} & \textbf{FFCL (Cortical Loop)}\\ \hline                        
1st Layer  & 46.36\%                                & 47.29\% &    47.32\%                      \\
2nd Layer & 46.56\%                                & 44.12\%  &   46.12\%                   \\
3rd Layer & 37.34\%                                & 43.69\% &    45.78\%                    \\
4th Layer  & 36.23\%                                & 42.86\% &   44.22\%                      \\
Model     & 47.78\% &          48.71\%                      &     49.93\%                     
\\ \hline
\end{tabular}
}
\caption{Comparative results of the Forward-Forward method and the New Ideas for CIFAR10 dataset}
\label{cifar10}
\vspace{-20pt}
\end{table}

Table \ref{Flops_count_table} details the computational complexity for a 4-layer model trained on the MNIST and Fashion MNIST datasets, as shown in Table \ref{model_architecture_table}. Table \ref{Flops_cifar} provides the inference complexity for the CIFAR 10 model. The FLOP counts are derived from equations \ref{FlopcountFFL} and \ref{FlopcountFFDL}.

\begin{table}[hbt!]
\centering
\scalebox{0.9}{
\begin{tabular}{lccc}
\hline
\textbf{Number of FLOPS}        & \textbf{FFL\cite{hinton2022forward}} & \textbf{FFDL} & \textbf{FFCL}\\ \hline                        
1st Layer & 420,000                                & 420,000                &  1,260,000      \\
2/3/4th Layers   & 2,510,000      & 306,000     &   918,000 \\
4 Layers Model &  7,950,000  & 1,338,000       &       4,014,000 \\
\hline                                                             
\end{tabular}
}
\caption{Comparative Results of FLOPS for the Forward-Forward Method vs. Label-Enhanced Hidden Layers Model on the MNIST and FashionMNIST Dataset, Using Models with 500 Neurons per Hidden Layer}
\label{Flops_count_table}
\vspace{-20pt}
\end{table}

By directly feeding labels, we can reuse neuron outputs to generate label-specific outputs in each layer without restarting the model for each label. This method significantly reduces the inference complexity in the 2nd, 3rd, and 4th layers, while maintaining the same complexity in the first layer. For instance, in models for MNIST and Fashion MNIST, total model complexity drops from 7.95M FLOPS to 1.34M FLOPS, a 5.9X reduction. While FFCL has a higher computational demand than FFDL due to executing the unrolled model, its complexity is still only half of FFL for MNIST and FashionMNIST, and almost a quarter for CIFAR 10. This shows that FFCL and FFDL not only improve accuracy over FFL but also greatly reduce inference complexity.

\begin{table}[hbt!]
\centering
\scalebox{0.9}{
\begin{tabular}{lccc}
\hline
\textbf{Number of FLOPS}        & \textbf{FFL\cite{hinton2022forward}} & \textbf{FFDL} & \textbf{FFCL}\\ \hline                        

1st Layer  &  62,955,520  & 6,520,832   &  19,562,496 \\
2nd Layer   & 20,992,000      & 2,211,840     &   6,635,520 \\
3rd Layer   & 5,253,120      & 581,632     &   1,744,896 \\
4th Layer   & 2,631,680      & 319,488     &   958,464 \\

4 Layers Model& 91,832,320       & 9,633,792           & 28,901,376 \\
\hline                                                             
\end{tabular}
}
\caption{Comparative Results of FLOPS for the Forward-Forward Method vs. Label-Enhanced Hidden Layers Model on the CIFAR10 Dataset, Using Models with 500 Neurons per Hidden Layer}
\label{Flops_cifar}
\vspace{-20pt}
\end{table}

\vspace{-10pt}
\section{Discussion on Future Work}
The FFL introduces a fresh approach to on-device learning by eliminating the conventional need for backpropagation. The FFDL and FFCL variants proposed in this paper further improve on FFL by increasing accuracy and reducing computational overhead. However, forward-forward learning is still in its early stages with many unanswered questions and many unexplored possibilities. In this section, we like to highlight some of the possibilities for extending the FFDL and FFCL. The FFCL uses the input to each unrolled copy of network. Introducing variations such as altered or time-series versions of the initial image could improve the model's robustness by enhancing the flow of information to each layer, potentially boosting resilience and performance. Another area for optimization is the FFCL's use of dense, inter-layer cortical feedback loops. Modifying these loops to extend across multiple layers or using them more sparingly could substantially simplify the current setup. Furthermore, changing the backpropagation requirements in FFL, FFCL, and FFDL reduces the need for backpropagation to a single layer, easing the storage burden of intermediate activations—a significant advantage for resource-limited edge devices. However, these models still require some backpropagation that typically depends on floating-point hardware. Researching ways to perform single-layer backpropagation on fixed-point hardware could further lessen memory requirements and eliminate the reliance on expensive floating-point computations.

\begin{acks}
This research was supported by the National Science Foundation under Award \#2203399.
\end{acks}

\vspace{-6pt}
\section{Conclusion}
In conclusion, this paper presented two novel variants of Forward-Forward Learning (FFL), each enhancing the framework in distinct ways to improve efficiency and accuracy. The first variant, Forward-Forward Direct Labeling (FFDL), incorporates direct label feeding into each voting layer, which significantly boosts model accuracy and reduces computational costs during inference by eliminating the need to rerun the entire network for different label votes. The second variant, Forward-Forward Cortical Loops (FFCL), builds on the direct label feeding strategy by integrating cortical loops, which allow for bidirectional information flow throughout the learning network, thereby further enhancing model accuracy.

\vspace{-6pt}
\bibliographystyle{ACM-Reference-Format}

\bibliography{ref}


\end{document}